\documentclass[10pt,twocolumn,letterpaper]{article}

\usepackage[pagenumbers]{cvpr} 
\usepackage[accsupp]{axessibility}
\usepackage{multirow}
\usepackage[table]{xcolor}

\definecolor{cvprblue}{rgb}{0.21,0.49,0.74}
\usepackage[pagebackref,breaklinks,colorlinks,allcolors=cvprblue]{hyperref}

\def\paperID{11023} 
\def\confName{CVPR}
\def\confYear{2025}

\title{Tora: Trajectory-oriented Diffusion Transformer for Video Generation}

\author{
    Zhenghao Zhang\textsuperscript{\rm 1}\footnotemark[1], ~~ Junchao Liao\textsuperscript{\rm 1}\footnotemark[1], ~~ Menghao Li\textsuperscript{\rm 1}, ~~ZuoZhuo Dai\textsuperscript{\rm 1},\\
    Bingxue Qiu\textsuperscript{\rm 1}, ~~ Siyu Zhu\textsuperscript{\rm 2}, ~~ Long Qin\textsuperscript{\rm 1}, ~~Weizhi Wang\textsuperscript{\rm 1}\\
    \textsuperscript{\rm 1} Alibaba Cloud Computing ~~~~~~ \textsuperscript{\rm 2} Fudan University \\
}

\begin{document}
\maketitle
\renewcommand{\thefootnote}{\fnsymbol{footnote}}
\footnotetext[1]{Equal contribution}
\renewcommand{\thefootnote}{\arabic{footnote}}

\begin{abstract}
Recent advancements in Diffusion Transformer (DiT) have demonstrated remarkable proficiency in producing high-quality video content. Nonetheless, the potential of transformer-based diffusion models for effectively generating videos with controllable motion remains an area of limited exploration. This paper introduces Tora, the first trajectory-oriented DiT framework that concurrently integrates textual, visual, and trajectory conditions, thereby enabling scalable video generation with effective motion guidance.
Specifically, Tora consists of a Trajectory Extractor~(TE), a Spatial-Temporal DiT, and a Motion-guidance Fuser~(MGF). The TE encodes arbitrary trajectories into hierarchical spacetime motion patches with a 3D motion compression network. The MGF integrates the motion patches into the DiT blocks to generate consistent videos that accurately follow designated trajectories.  Our design aligns seamlessly with DiT's scalability, allowing precise control of video content's dynamics with diverse durations, aspect ratios, and resolutions. Extensive experiments demonstrate that Tora excels in achieving high motion fidelity compared to the foundational DiT model, while also accurately simulating the complex movements of the physical world.
Code is made available at \href{https://github.com/alibaba/Tora}{https://github.com/alibaba/Tora} .
\end{abstract}    
\section{Introduction}

\begin{figure*}[!t]
    \centering
    \includegraphics[width=0.94\textwidth]{images/f2.pdf}
    \caption{
         More generated samples. Tora effectively manages trajectories to precisely manipulate various objects and backgrounds. In the realm of image-to-video synthesis, it can craft dynamic camera movements in accordance with textual descriptions and the designated starting trajectory points, such as common backgrounds.  Furthermore, Tora supports video generation across different aspect ratios, resolutions, and durations, ensuring flexible content creation. 
    }
    \label{fig:2}   
\end{figure*}

Diffusion models~\cite{Dhariwal2021DiffusionMB, ramesh2022hierarchical} have demonstrated their capability to generate diverse and high-quality images or videos. Previously, video diffusion models~\cite{ho2022video, DBLP:journals/corr/abs-2311-15127,DBLP:journals/corr/abs-2311-04145} predominantly employed UNet-based architectures~\cite{ronneberger2015u}, focusing primarily on synthesizing videos of limited duration, typically around two seconds, and were constrained to fixed resolutions and aspect ratios. Recently, Sora~\cite{sora2024}, a text-to-video generation model leveraging Diffusion Transformer~(DiT)~\cite{peebles2023scalable}, has showcased video generation capabilities that significantly outstrip current state-of-the-art methods. Sora excels not only in the production of high-quality videos ranging from 10 to 60 seconds, but also distinguishes itself through its capacity to handle diverse resolutions, various aspect ratios, adherence to the laws of actual physics.

Video generation requires consistent motion across image sequences, underscoring the importance of motion control. Previous works, such as VideoComposer~\cite{wang2023videocomposer} and DragNUWA~\cite{yin2023dragnuwa}, have implemented generalized motion manipulation through motion vectors and trajectories. Building on this foundation, MotionCtrl~\cite{wang2024motionctrl} innovates by independently managing camera and object motions, thereby expanding the diversity of achievable motion patterns. Despite their promising controllable motion quality, UNet-based methods are restricted to generating videos of only 16 frames at a fixed, lower resolution. This limitation hinders the smooth portrayal of motion, particularly during significant positional shifts in the provided trajectory, leading to distortion and unnatural movements, such as parallel drifting, which diverge from real-world dynamics. Consequently, there is an urgent need for a model capable of producing longer videos with robust motion control and detailed physical representations.

To address these challenges, we present Tora, the first DiT model that simultaneously integrates text, images, and trajectories, enabling scalable video generation with robust motion control.
Notably, our work adopts OpenSora~\cite{OpenSora}, an open-source version of Sora, as the foundational DiT model. To align motion control with the scalability of the DiT framework, we propose two novel modules: the Trajectory Extractor (TE), which converts arbitrary trajectories into hierarchical spacetime motion patches, and the Motion-guidance Fuser (MGF), designed to seamlessly integrate these patches within the stacked DiT blocks.
More specifically, TE initially converts positional displacements along trajectory into the RGB domain via flow visualization techniques. These visualized displacements undergo Gaussian filtering to mitigate scattered issues. Subsequently, a 3D Variational Autoencoder (VAE)~\cite{kingma2013auto} encodes trajectory visualizations into spacetime motion latents, which share the same latent space with video patches. The motion latents are then decomposed into multiple levels of motion conditions via stacked lightweight modules. Our VAE architecture is inspired by MAGVIT-v2~\cite{DBLP:journals/corr/abs-2310-05737} but simplified by omitting codebook dependencies. The MGF integrates adaptive normalization layers~\cite{DBLP:conf/aaai/PerezSVDC18} to infuse multi-level motion conditions into the corresponding DiT blocks. We explored various adaptations of transformer blocks including adaptive layer normalization, cross-attention, and extra channel connections to inject the motion conditions. Among these, adaptive layer normalization emerged as the most effective to generate consistent videos following the trajectory.

During training, we adapt OpenSora's workflow to generate high-quality video-text pairs and utilize an optical flow estimator~\cite{DBLP:journals/pami/XuZCRYTG23} for trajectory extraction. We also integrate a motion segmentor~\cite{DBLP:conf/eccv/ZhaoLGWL22} with a camera detector\footnote{https://github.com/antiboredom/camera-motion-detector} to filter out instances dominated by camera motion, thereby improving our tracking of object trajectories. This careful selection process results in a dataset of high-quality videos with consistent motion. With an adapter-like strategy~\cite{DBLP:conf/aaai/MouWXW0QS24}, we solely train the temporal blocks, together with the TE and MGF. This strategy seamlessly integrates DiT's inherent generative knowledge with external motion signals.


The main contributions of our work are as follows:

\begin{itemize}
\item 
We introduce Tora, the first trajectory-oriented DiT model for video generation. As illustrated in Figure~\ref{fig:2}, Tora enables the creation of motion-manipulable videos with varying aspect ratios, extending up to 204 frames and 720p resolution.

\item 
We propose a novel trajectory extractor and a motion-guidance fusion mechanism to facilitate motion control that aligns with the scalability of DiT. Additionally, we ablate several architectural choices and scaling capabilities to offer empirical baselines for future research.

\item 
Experiments demonstrate that Tora achieves state-of-the-art accuracy in controlling object motions. Furthermore, it demonstrates superiority in simulating movements within the physical world.
\end{itemize}

\section{Related Work}
\subsection{Diffusion models for Video Generation}
Diffusion models have demonstrated an impressive capability to generate high-quality video samples. Previous research~\cite{ho2022video, ho2022imagen, singer2022make,khachatryan2023text2video,zhang2023controlvideo} commonly use video diffusion models~(VDMs) that incorporate temporal convolutional and attention layers into the pre-trained image diffusion models. Subsequently, VideoCrafter~\cite{Chen2023VideoCrafter1OD} and SVD~\cite{DBLP:journals/corr/abs-2311-15127} expand the application of video diffusion models to larger datasets, while TF-T2V~\cite{wang2024tf} directly learns from extensive text-free videos. Nonetheless, these methods encounter limitations in generating long videos, owing to the inherent constraints on capacity and scalability within the UNet design. Conversely, DiT-based models~\cite{sora2024, OpenSora,DBLP:journals/corr/abs-2405-04233, yang2024cogvideox, polyak2024movie} can directly generate videos extending up to tens of seconds. Sora~\cite{sora2024} converts visual data into a unified representation, facilitating large-scale training and enabling the generation of 1-minute high-definition video. Vidu~\cite{DBLP:journals/corr/abs-2405-04233} is capable of generating both realistic and imaginative videos in various aspect ratios. CogVideoX~\cite{yang2024cogvideox}  introduces an expert diffusion transformer model that generates videos from text prompts or images, along with an effective text-video data processing pipeline that enhances video caption quality. 

\begin{figure*}[!t]
    \centering
    \includegraphics[width=0.95\textwidth]{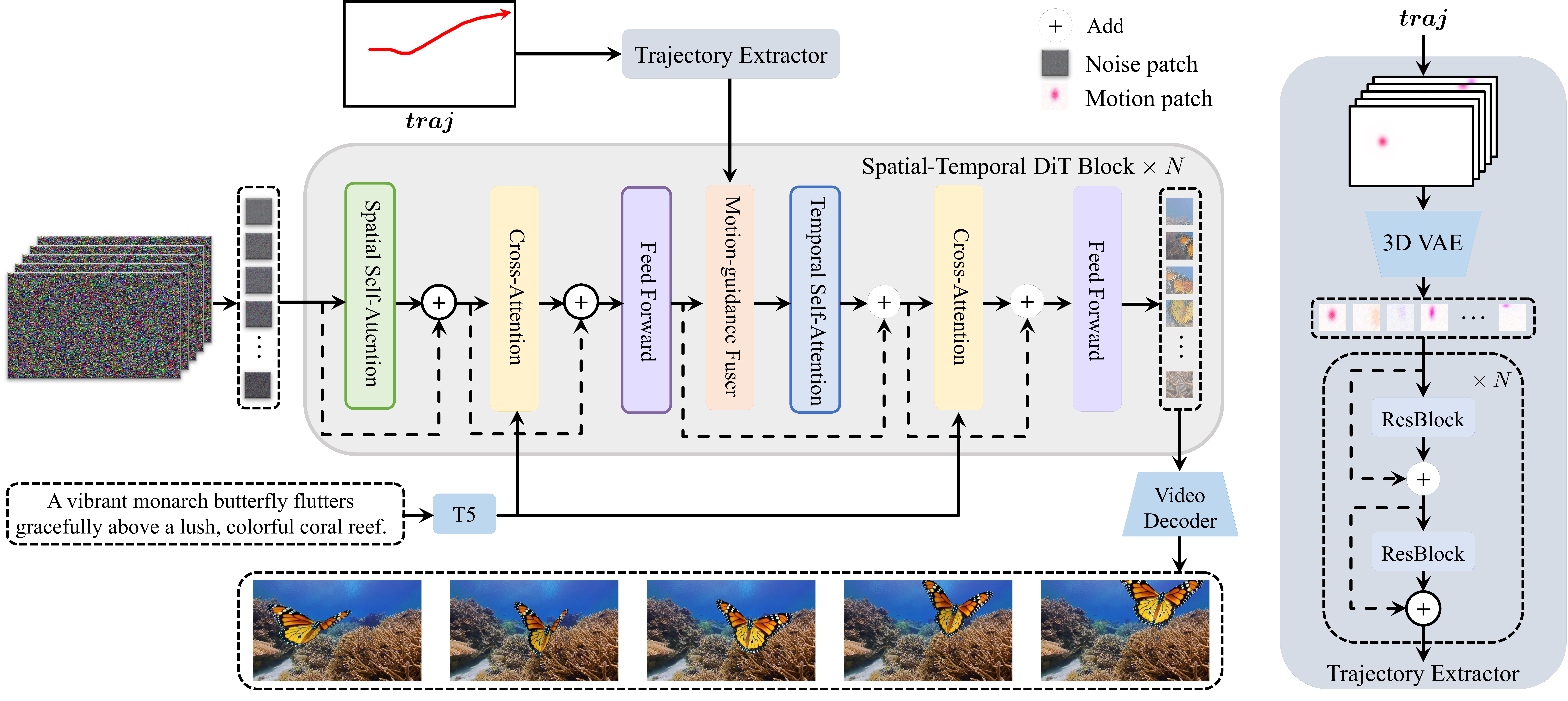}
    \caption{
    Overview of the Tora Architecture. We introduce two novel modules: the Trajectory Extractor and the Motion-guidance Fuser. The Trajectory Extractor uses a 3D motion VAE to embed trajectory vectors into the same latent space as video patches, preserving motion information across frames. It then employs stacked convolutional layers to extract hierarchical motion features. The Motion-guidance Fuser utilizes adaptive normalization layers to integrate these multi-level motion conditions into the corresponding DiT blocks, ensuring that generated videos consistently follow defined trajectories. Our method leverages the scalability of DiT, enabling the creation of motion-controllable videos of extended duration.
    }
    \label{fig:3}   
\end{figure*}

\subsection{Motion control in Video Generation}
To better control motion in generated videos,  a multitude of studies have endeavored to introduce diverse motion signals in VDMs.
Pioneering works like MotionDirector~\cite{DBLP:journals/corr/abs-2310-08465} and VMC~\cite{jeong2024vmc} have utilized reference videos to extract motion patterns applicable to diverse video generations. VideoComposer~\cite{wang2023videocomposer} expands upon this by adopting depth maps, sketches, or motion vectors from references as conditional inputs for motion control. Nonetheless, these methodologies are limited to reproducing existing motion patterns. Conversely, approaches that leverage trajectories~\cite{mou2024revideoremakevideomotion,geng2024motionpromptingcontrollingvideo,shi2024motioni2vconsistentcontrollableimagetovideo,yin2023dragnuwa,wang2024motionctrl,DBLP:journals/corr/abs-2401-00896} or bounding boxes~\cite{yin2023dragnuwa,DBLP:journals/corr/abs-2311-12886, wang2024motionctrl} in video generation promise greater adaptability and user accessibility.
DragNUWA~\cite{yin2023dragnuwa} breaks new ground by integrating trajectory-based conditioning into VDMs, facilitating complex camera and object movements. AnimateAnything~\cite{DBLP:journals/corr/abs-2311-12886} employs motion masks for precise control over the moving regions. DragAnything~\cite{wu2024draganythingmotioncontrolusing} uses the object mask to generate entity representations for achieving motion control. 
MotionCtrl~\cite{wang2024motionctrl} facilitates more flexible control, allowing separate adjustment of both camera movements and individual object motions.
However, all of them yield noticeable artifacts in both motion consistency and visual presentation when applied to longer sequences.  In contrast, our method first integrates trajectories into DiT architecture, which enables closer adherence to the physical world.

\section{Methodology}
\subsection{Preliminary}
\textbf{Latent Video Diffusion Model~(LVDM).} The LVDM enhances the stable diffusion model~\cite{ramesh2022hierarchical} by integrating a 3D UNet, thereby empowering efficient video data processing. This 3D UNet design augments each spatial convolution with an additional temporal convolution and follows each spatial attention block with a corresponding temporal attention block. It is optimized by employing a noise-prediction objective function:
\begin{equation}
    l_\epsilon = ||\epsilon - \epsilon_\theta(z_t, t, c)||^2_2,
    \label{eq:training_objective}
\end{equation}
Here, $\epsilon_\theta(\cdot)$ signifies the 3D UNet's noise prediction function. The condition 
$c$ is guided into the UNet using cross-attention for adjustment. Meanwhile, $z_t$ denotes the noisy hidden state, evolving like a Markov chain that progressively adds Gaussian noise to the initial latent state $z_0$:
\begin{equation}
    z_t=\sqrt{\bar{\alpha}_t}z_0 + \sqrt{1 - \bar{\alpha}_t}\epsilon, \quad\epsilon \sim \mathcal{N}(0, I),
    \label{eq:add_noise}
\end{equation}
where $\bar{\alpha}_t = \prod_{i=1}^t(1-\beta_t)$ and $\beta_t$ is a coefficient that controls the noise strength in step $t$.

\noindent \textbf{Diffusion Transformer~(DiT).} The DiT~\cite{peebles2023scalable} introduces a novel architecture that merges the strengths of diffusion models with transformer architectures~\cite{DBLP:conf/nips/VaswaniSPUJGKP17}.  This integration aims to address the limitations of traditional UNet-based latent diffusion models (LDMs), improving their performance, versatility, and scalability. While keeping the overall framework consistent with existing LDMs, the key shift lies in replacing the UNet with a transformer architecture for learning the denoising function $\epsilon_\theta(\cdot)$, thereby marking a pivotal advance in the realm of generative modeling.


\subsection{Tora}
Tora employs the Spatial-Temporal Diffusion Transformer (ST-DiT) from OpenSora as its foundational model. To facilitate user-friendly motion control while aligning with the scalability of DiT, Tora integrates two novel motion-processing components: the Trajectory Extractor (TE) and the Motion-guidance Fuser (MGF). An overview of Tora's workflow is illustrated in Figure~\ref{fig:3}.

\noindent \textbf{Spatial-Temporal DiT.} The ST-DiT architecture incorporates two distinct block types: the Spatial DiT Block (S-DiT-B) and the Temporal DiT Block (T-DiT-B), arranged in an alternating sequence. The S-DiT-B comprises two attention layers, each performing Spatial Self-Attention (SSA) and Cross-Attention sequentially, succeeded by a point-wise feed-forward layer that serves to connect adjacent T-DiT-B block. Notably, the T-DiT-B modifies this schema solely by substituting SSA with Temporal Self-Attention (TSA), preserving architectural coherence. Within each block, the input, upon undergoing normalization, is concatenated back to the block's output via skip-connections. By leveraging the ability to process variable-length sequences, the denoising ST-DiT can handle videos of variable durations.

During processing, a video autoencoder~\cite{yu2023magvit} is first employed to diminish both spatial and temporal dimensions of videos. To elaborate, it encodes the input video $X \in \mathbb{R}^{L \times H \times W \times 3}$ into video latent $z_{0} \in \mathbb{R}^{l \times h \times w \times 4}$, where $L$ denotes the video length and $l = L / 4, h = H / 8, w = W / 8$.  $z_{0}$ is next ``patchified", resulting in a sequence of input tokens $I \in \mathbb{R}^{l \times s \times d} $. Here, $s = hw/p^2$ and $p$ denotes the patch size. 
In both SSA and TSA, standard Attention is performed using Query ($Q$), Key ($K$), and Value ($V$) matrices:
\vspace{-2mm}
\begin{equation}
Q = W_{Q} \cdot I_\mathrm{norm}; K = W_{K} \cdot I_\mathrm{norm}; V = W_{V} \cdot I_\mathrm{norm},
\end{equation}
Here, $I_\mathrm{norm}$ is the normalized $I$, $W_{Q},W_{K}, W_{V}$ are learnable matrices.
The textual prompt is embedded with a T5 encoder and integrated using a cross-attention mechanism.

\noindent \textbf{Trajectory Extractor.} Trajectories have proven to be a more user-friendly method for controlling the motion of generated videos. Specifically, given a trajectory $traj=\left \{ (x_i, y_i) \right \} _{i=0}^{L-1} $, where $(x_i, y_i)$ denotes the spatial position $(x, y)$ at the $i$-$th$ frame the trajectory passes through. Previous studies primarily encode the horizontal offset $u(x_i,y_i)$ and the vertical offset $v(x_i,y_i)$ as the motion condition: 
\begin{equation}
u(x_i,y_i)= x_{i+1} - x_{i}; ~ v(x_i,y_i)= y_{i+1} - y_{i},
\end{equation}
However, the DiT model employs a video autoencoder and a patchification process to convert the video into patches. Here, each patch is derived across multiple frames, rendering it inappropriate to straightforwardly employ frame-to-frame offsets. To address this, our TE converts the trajectory into motion patches, which inhabit the same latent space as the video patches. Particularly, we begin by transforming the $traj$ into a trajectory map $g \in \mathbb{R}^{L \times H \times W \times 2}$, enhanced with a Gaussian Filter to mitigate scatter. Notably, the first frame employs a fully-zero map. Afterward, the trajectory map $g$ is translated into the RGB color space, producing $g_{vis} \in \mathbb{R}^{L \times H \times W \times 3}$ through a flow visualization technique. We use a 3D VAE to compress trajectory maps, achieving an 8x spatial and 4x temporal reduction, aligning with OpenSora framework. Our VAE is based on the MAGVIT-v2 architecture, with spatial compression initialized using the VAE of SDXL~\cite{DBLP:journals/corr/abs-2307-01952} to accelerate convergence. We train the model using reconstruction loss to obtain the compact motion latent representation $g_{m} \in \mathbb{R}^{l \times h \times w \times 4}$ from the $g_{vis}$.

To match the size of the video patches, we use the same patch size on 
$g_{m}$ and encode it using a series of convolutional layers, resulting in spacetime motion patches $f \in \mathbb{R}^{l \times s\times d' }$. Here $d'$ is the dimension of motion patches. The output of each convolutional layer is skip-connected to the input of the next layer to extract multi-level motion features:
\begin{equation}
f_i= \mathrm{Conv}^{i}~(f_{i - 1}) + f_{i - 1},
\end{equation}
where $f_i$ is the motion condition for $i$-$th$ ST-DiT block.

\begin{figure}[!t]
    \centering
    \includegraphics[width=0.47\textwidth]{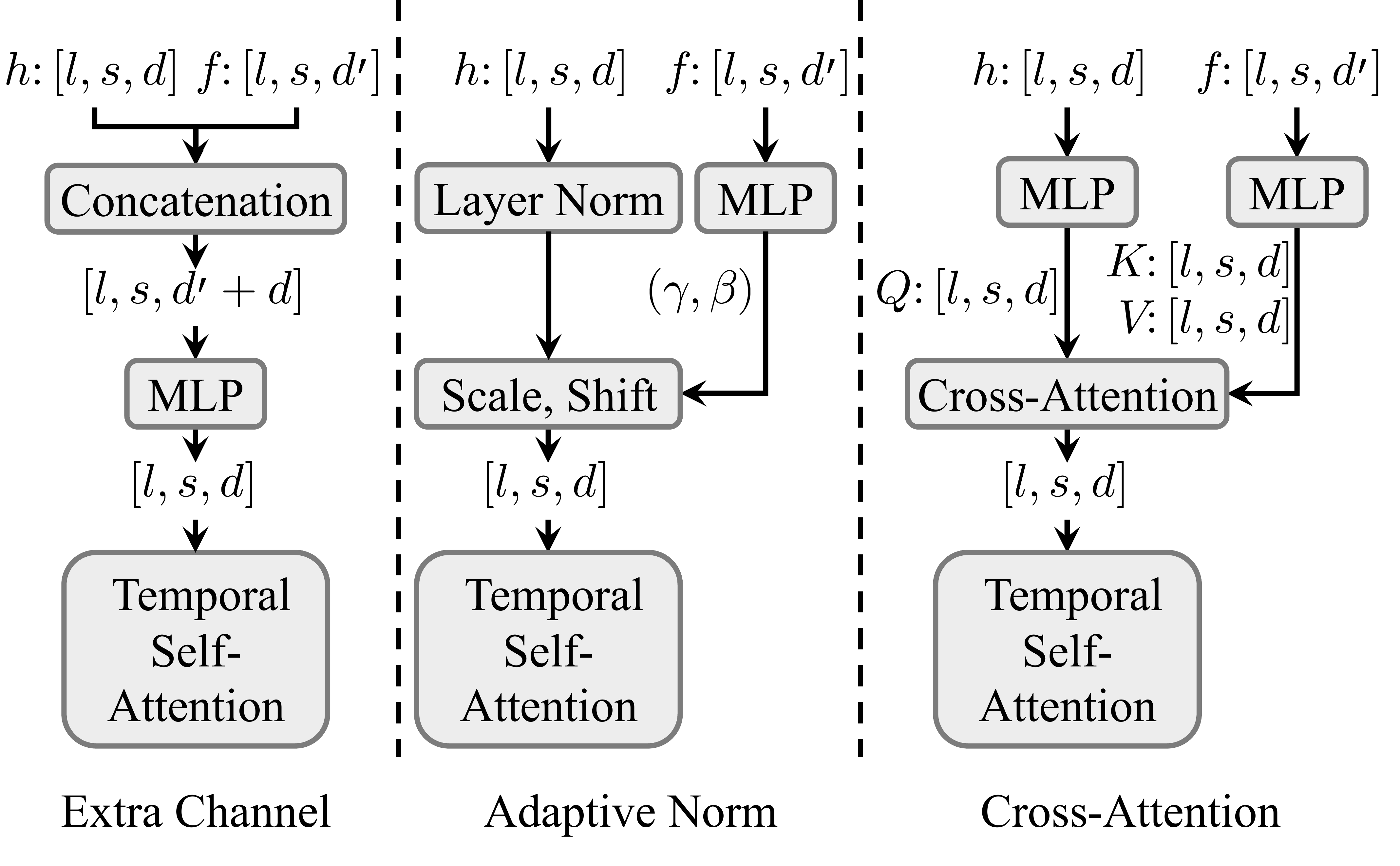}
    \caption{
        Different designs of the Motion-guidance Fuser for incorporating trajectory conditioning. Adaptive Norm demonstrates the best performance.
    }
    \label{fig:4}   
\end{figure}

\noindent \textbf{Motion-guidance Fuser.} To incorporate DiT-based video generation with the trajectory, we explore three variants of fusion architectures that inject motion patches into each ST-DiT block. These designs are illustrated in Figure~\ref{fig:4}.

\begin{itemize}[label=-]
\item Extra channel connections. Denote $h_i \in \mathbb{R}^{l \times s \times d} $ as the resultant output from the $i$-$th$ block of the ST-DiT. Following the widespread use of concatenation in GAN-based LVDM, the motion patches are simply concatenated with the previous hidden state $h_{i-1}$ along the channel dimension. An additional MLP is then added to maintain the same latent size:
\begin{equation}
h_{i} = \mathrm{MLP}~([h_{i-1}, f_i]) + h_{i-1},
\end{equation}
\item Adaptive Norm layer. Inspired by the adaptive normalization layers employed in GANs, we initially convert $f_i$ into scale $\gamma_{i}$ and shift $\beta_{i}$ by adding two zero-initialized convolution layers into the ST-DiT block. Subsequently, $\gamma_{i}$ and $\beta_{i}$ are integrated into $h_i$ through a straightforward linear projection:
\begin{equation}
h_{i} = \gamma_{i} \cdot h_{i-1} + \beta_{i} + h_{i-1},
\end{equation}

\item Cross-Attention layer. The ST-DiT block has been modified to include an additional Cross-Attention layer following the SSA or TSA, with the motion patches serving as the key and value to integrate with the hidden state h: 

\begin{equation}
h_{i} = \mathrm{CrossAttn}~([h_{i-1}, f_i]) + h_{i-1},
\end{equation}
\end{itemize}


We evaluate three types of fusion architectures and find that the adaptive norm yields the best performance and computational efficiency. For the remainder of the paper, MGF employs the adaptive norm layer unless otherwise specified.

\subsection{Data Processing and Training Strategy}

\noindent \textbf{Data Processing}. We employ a structured data processing method to obtain high-quality training videos with consistent object motion. Initially, raw videos are segmented into shorter clips based on scene detection\footnote{https://github.com/Breakthrough/PySceneDetect}. Subsequently, we remove invalid videos, such as those with encoding errors, zero duration, or low resolution. Furthermore, we utilize aesthetic\footnote{https://github.com/christophschuhmann/improved-aesthetic-predictor} and optical flow scores~\cite{DBLP:journals/pami/XuZCRYTG23} to filter out low-quality videos. To concentrate on the motion of primary objects, we implement camera motion filtering, using results from motion segmentation~\cite{DBLP:conf/eccv/ZhaoLGWL22} and camera detection to exclude instances predominantly exhibiting camera movement. Dramatic object motions in certain videos can lead to significant optical flow deviations, which may interfere with trajectory training. To address this, we retain these videos based on a probability of $(1 - flow\_score / 100)$.  For eligible videos, we generate captions using the PLLaVA model~\cite{DBLP:journals/corr/abs-2404-16994}. During inference, we utilize GPT-4o~\cite{DBLP:journals/corr/abs-2303-08774} to refine prompts, ensuring alignment with training process. 

\noindent \textbf{Motion condition training}. Inspired by DragNUWA and MotionCtrl, we adopt a two-stage training approach for trajectory learning. In the first stage, we extract dense optical flow \cite{DBLP:journals/pami/XuZCRYTG23} from the training video as the trajectory, providing richer information to enhance motion learning. In the second stage, we adjust the model from complete optical flow to more user-friendly trajectories by randomly selecting 1 to $N$  object trajectories based on motion segmentation results and flow scores. To improve the scattered nature of sparse trajectories, we apply a Gaussian filter for refinement. After completing the two-stage training, Tora facilitates flexible motion control using arbitrary trajectories.

\noindent \textbf{Image condition training}. We employ a mask strategy to support visual conditioning. Specifically, we randomly unmask frames during training, ensuring that the video patches of unmasked frames are not subjected to any noise.


\section{Experiments}

\subsection{Experimental Setup}
\noindent \textbf{Implementation Details.} Tora is initialized with OpenSora v1.2 weights, and training videos have resolutions from 144p to 720p and frame counts ranging from 51 to 204. To balance training FLOP and memory usage, we adjust the batch size from 1 to 50. We use Adam Optimizer~\cite{DBLP:journals/corr/KingmaB14} with a learning rate of $2\times10^{-5}$ on 8 NVIDIA A100. The 3D VAE is initially trained on datasets ~\cite{DBLP:conf/cvpr/MehlSJNB23,DBLP:conf/cvpr/MayerIHFCDB16,DBLP:journals/ijcv/RanjanHTTRB20,DBLP:journals/corr/abs-2001-10773} for optical flow estimation and then frozen during Tora training. We train Tora for 2 epochs with dense optical flow and fine-tune for 1 epoch with sparse trajectories. The maximum number of sampling trajectories $N$ is set to 16. The inference step and the guidance scale are set to 30 and 7.0, respectively.

\noindent \textbf{Dataset.} Our training videos are sourced from four datasets: 1) Panda-70M~\cite{DBLP:journals/corr/abs-2402-19479}, from which we use the training-10M subset containing high-quality videos; 2) Mixkit~\cite{mixkit}; 3) Pexels~\cite{pexel}; and 4) Internal videos. The internal videos are manually annotated, with each clip labeled to include object masks and camera movement. Following our data processing pipeline, we select about 630k eligible videos for the training dataset. For inference, we curate 185 clips with diverse motion trajectories and scenes, to serve as a new benchmark for evaluating the motion controllability. 

\noindent \textbf{Metrics.} We leverage standard metrics including Fréchet Video Distance (FVD)~\cite{Thomas2018fvd}, and CLIP Similarity (CLIPSIM)~\cite{wu2021godiva} to quantitatively evaluate video quality. 
For assessing motion controllability, we utilize the Trajectory Error (TrajError) metric, which calculates the average L1 distance between the generated and predefined trajectories. 

\subsection{Results}

\begin{table*}[!t]\small
\setlength{\tabcolsep}{3pt}
\centering
\begin{tabular}{ccccccccccccc}
\toprule
\multirow{2}{*}{Method} & \multicolumn{3}{c}{FVD~($\downarrow$)} & \multicolumn{3}{c}{CLIPSIM~($\uparrow$)} & \multicolumn{3}{c}{TrajError~($\downarrow$)} \\ \cmidrule(lr){2-4} \cmidrule(lr){5-7} \cmidrule(lr){8-10}
 & 16-frame & 64-frame & 128-frame & 16-frame & 64-frame & 128-frame & 16-frame & 64-frame & 128-frame \\ 
\midrule
\rowcolor{black!10} \multicolumn{1}{l}{\textbf{\textit{UNet-based method}}} &&&&&&&&& \\
VideoComposer~\cite{wang2023videocomposer} & 529 & 668 & 856 & 0.2335 & 0.2284 & 0.2236 & 15.11 & 29.14 & 58.76    \\
DragNUWA~\cite{yin2023dragnuwa} & 475 & 593 & 784 & 0.2385 & 0.2341 & 0.2305 & 10.04 & 17.33 & 41.25    \\
AnimateAnything~\cite{DBLP:journals/corr/abs-2311-12886} & 487 & 602 & 775 & 0.2399 & 0.2342 & 0.2313 & 13.39 & 27.28 & 51.33     \\
TrailBlazer~\cite{DBLP:journals/corr/abs-2401-00896} & 459 & 581 & 756 & 0.2403 & 0.2351 & 0.2322 & 11.68 & 19.47 & 44.10    \\
MotionCtrl~\cite{wang2024motionctrl} & 463 & 572 & 731 & 0.2412 & 0.2376 & 0.2331 & 9.42 & 16.46 & 38.39  \\
\midrule
\rowcolor{black!10} \multicolumn{1}{l}{\textbf{\textit{DiT-based method}}} &&&&&&&&&\\
OpenSora~\cite{OpenSora} & \textbf{430} & 476 & 533  & \textbf{0.2452} & 0.2433 & 0.2411 & 286.43 & 321.52 & 373.17         \\
OpenSora-based DragNUWA*  & 451 & 504 & 565  & 0.2430 & 0.2419 & 0.2393 & 10.11 & 13.88 & 21.75     \\
\textbf{Tora(Ours)} & 438 & \textbf{460} & \textbf{494} & 0.2447 & \textbf{0.2435} & \textbf{0.2418} & \textbf{7.23} & \textbf{8.45} & \textbf{11.72}     \\ 
\bottomrule
\end{tabular}
\caption{
Quantitative comparisons with motion-controllable video generation models. As the number of generated frames
increases, Tora's performance advantage over UNet-based methods becomes more pronounced. Specifically, Tora not only enhances motion fidelity but also improves the visual quality of the foundational model. Comparisons with OpenSora-based DragNUWA highlight the strengths of our proposed motion modules, which integrate seamlessly with DiT’s architecture.}
\label{tab1}
\vspace{-1mm}
\end{table*}


We evaluate against motion-guided video generation methods using 16-frame, 64-frame, and 128-frame configurations. Trajectories are scaled proportionally for different durations. UNet-based approaches utilize sequential inference for extended generation. Since there are no DiT-based methods, we adapt DragNUWA's motion trajectory design to our foundation model as an additional baseline. Specifically, we implement its official convolutional motion feature extraction and linear projection-based injection. To ensure compatibility with base DiT model, we perform downsampling across both spatial and temporal dimensions.

\begin{figure*}[!t]
    \centering
    \includegraphics[width=0.95\textwidth]{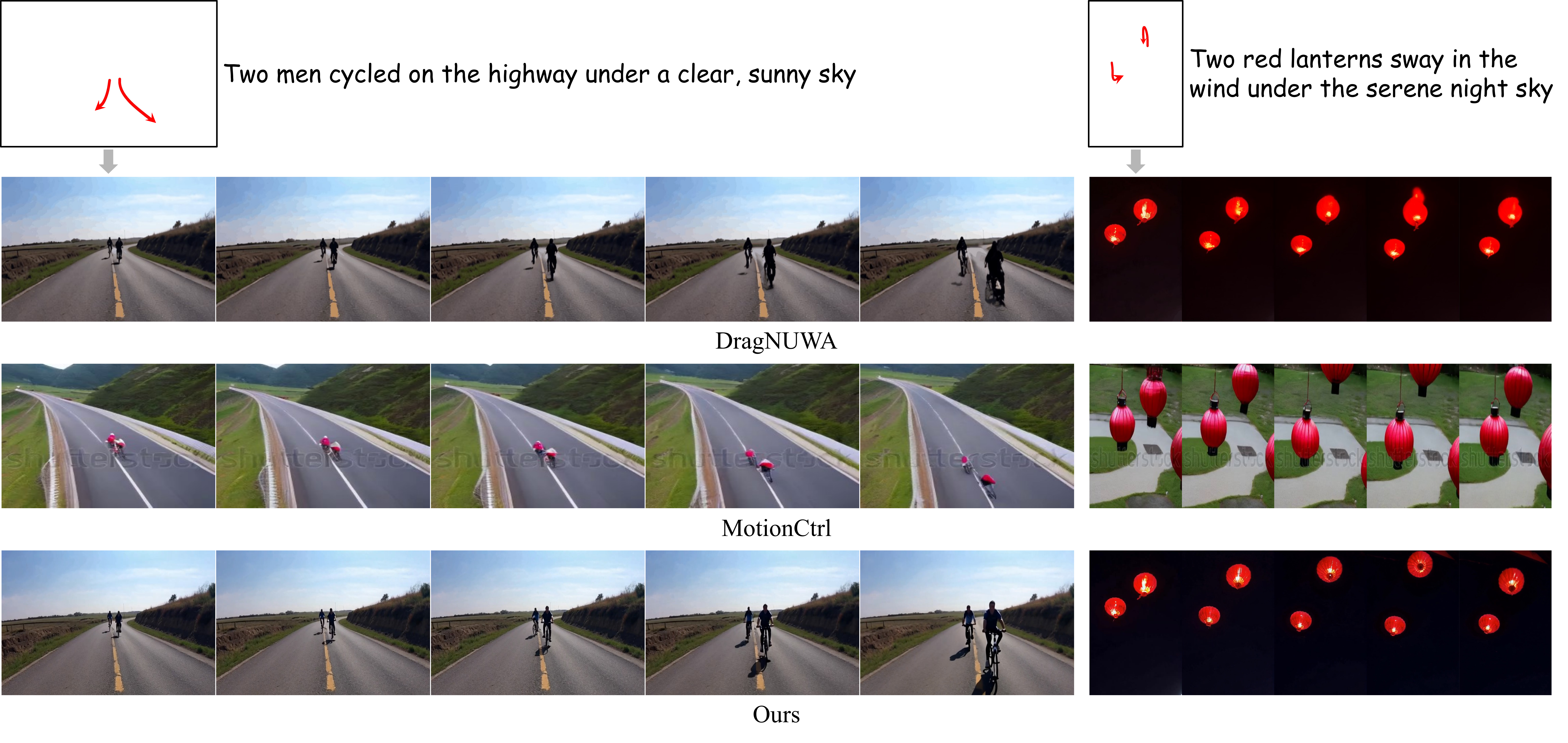}
    \caption{
        Qualitative Comparisons on Trajectory Control. 
        In the bicycle scenario, Tora realistically captures pedaling motions, while other methods show legs in an unnatural, nearly horizontal position. In another case, DragNUWA causes significant deformation of the lanterns, and MotionCtrl fails to accurately depict two lanterns. Overall, Tora not only adheres precisely to the specified trajectory but also produces smoother movement that conforms to the physical world. 
    }
    \label{fig:6}   
    \vspace{-2mm}
\end{figure*}

As shown in Table~\ref{tab1}, UNet-based methods all exhibit increasing deviations with longer sequences, causing motion blur and object deformation. In contrast, Tora leverages the transformer's scaling capabilities to maintain robustness, achieving 3-5 times higher trajectory accuracy and approximately 30-40$\%$ better FVD in 128-frame tests. Figure~\ref{fig:6} shows a comparative qualitative analysis of Tora against these mainstream motion control techniques.

While the OpenSora achieves high visual quality, its inability to incorporate motion control leads to random object trajectories, resulting in significantly elevated TrajError. Notably, Tora demonstrates dual superiority over OpenSora in both motion control efficacy and visual performance at most settings. We find Tora's exceptional capacity in suppressing temporal artifacts and motion blurring compared with OpenSora (see Appendix for visual comparisons), yielding enhanced temporal coherence and video stability. While OpenSora-based DragNUWA improves motion controllability over UNet-based methods, its motion representations exhibit intrinsic incompatibility with DiT's latent space. This architectural incompatibility induces suboptimal feature integration during training, consequently degrading visual quality by about 5$\%$ below baseline. 

\begin{figure}[!t]
    \centering
    \includegraphics[width=0.45\textwidth]{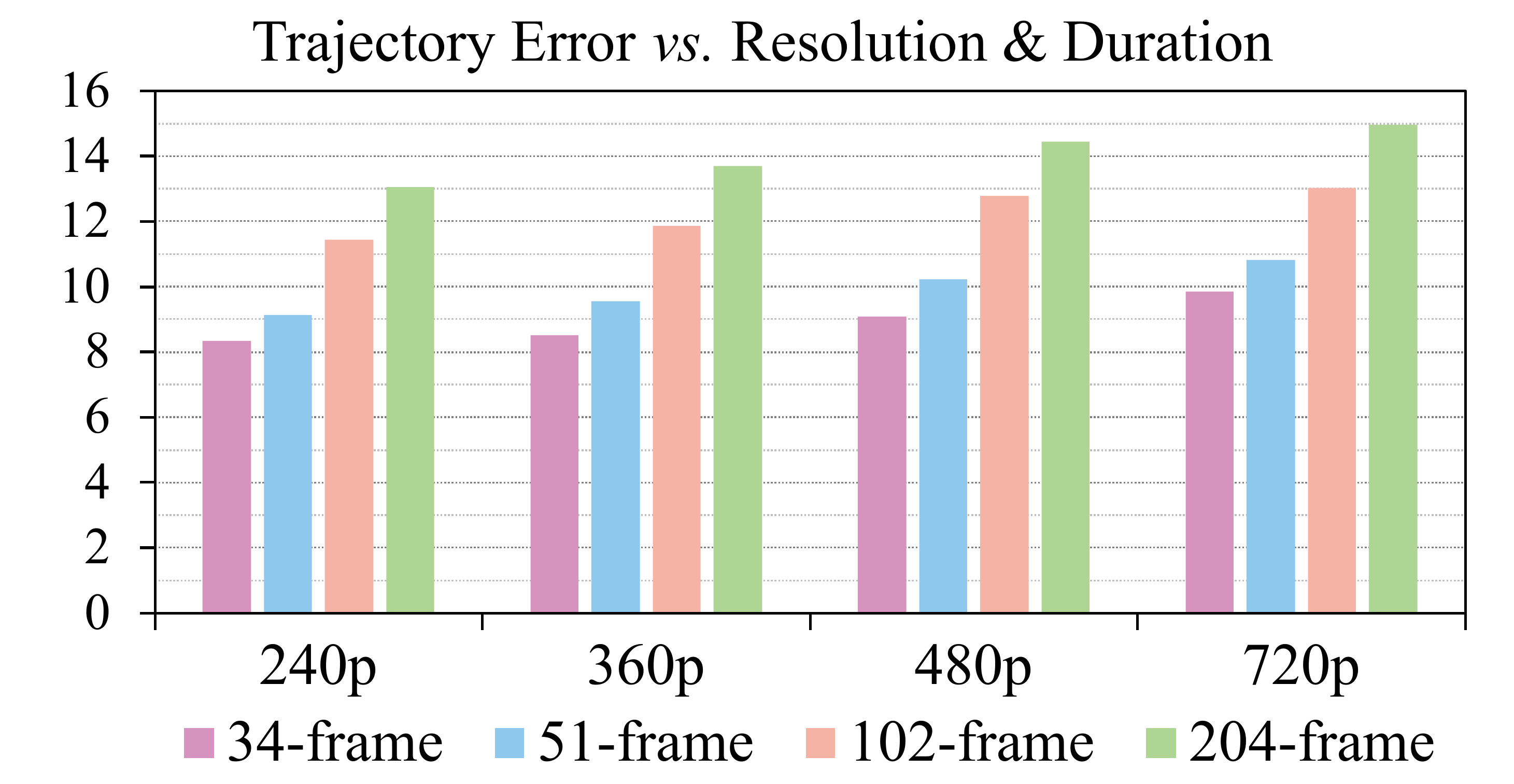}
    \caption{
        Comparison of Trajectory Error across various resolutions and durations. 
    }
    \label{fig:5}   
\end{figure}

Figure~\ref{fig:5} presents an analysis of Trajectory Error across different resolutions and durations. Unlike UNet models, Tora shows a gradual increase in error as duration extends. This aligns with the decrease in video quality observed in the DiT model. The results demonstrate that our method effectively maintains trajectory control over longer durations.

\begin{table}[!t]\footnotesize
\centering
\renewcommand{\arraystretch}{1.1}
\begin{tabular}{cccc}
    \toprule
    Method             & FVD~($\downarrow$) & CLIPSIM~($\uparrow$) & TrajError~($\downarrow$) \\ \midrule
    Sampling Frame &  581   &  0.2304       &    27.61              \\
    Average Pooling    &  558   &  0.2325    &          20.97       \\
    3D VAE             &  \textbf{513}   &  \textbf{0.2358}       &   \textbf{14.25}               \\ 
    \bottomrule
\end{tabular}
\caption{Evaluation of the impact of different trajectory compression methods.} 
\vspace{-3mm}
\label{tab2}
\end{table}

\subsection{Ablation study}

\noindent \textbf{Trajectory Compression.} To validate the alignment of motion patches with the DiT inputs, we explore three different methods for trajectory compression, as summarized in Table~\ref{tab2}. The first method samples the mid-frame as a keyframe for successive 4-frame intervals and uses Patch-Unshuffle~\cite{jang2023pucapatchunshufflechannelattention} for spatial compression. While simple, this approach is sub-optimal for motion control due to potential flow estimation errors during rapid movements or occlusions, and the increased dissimilarity between patches complicates learning. The second method employs average pooling to gather information from successive frames. Although this captures a general sense of movement, it sacrifices precision by averaging the trajectory's direction and magnitude, diluting important motion details. Our method utilizes a 3D motion VAE to extract the global context of successive trajectory intervals. 
Extensive training on a large dataset of trajectory videos with this method yields the best results, highlighting the effectiveness of our customized VAE approach for compression.


\begin{table}[!t]\footnotesize
\centering
\renewcommand{\arraystretch}{1.1}
\begin{tabular}{cccc}
\toprule
Method             & FVD~($\downarrow$) & CLIPSIM~($\uparrow$) & TrajError~($\downarrow$) \\ \midrule
Extra Channel &   542  &    0.2329     &   21.07               \\
Cross Attention    &   526  &    0.2354     &   18.36               \\
Adaptive Norm        & \textbf{513}    &   \textbf{0.2358}     & \textbf{14.25}                 \\ 
\bottomrule
\end{tabular}
\caption{Different variants of motion fusion blocks employed in MGF. Adaptive Norm works best.}
\label{tab3}
\vspace{-3mm}
\end{table}

\noindent \textbf{Block design and integrated position of MGF.} We train the three variant MGF blocks as previously described, with the results presented in Table~\ref{tab3}. Notably, the adaptive norm block achieves lowest FVD and Trajectory Error, while also exhibiting the highest computational efficiency. This advantage arises from its ability to dynamically adapt features based on varying conditions without strict alignment, a common cross-attention challenge. It also maintains temporal consistency by modulating conditional information over time, essential for incorporating motion cues. In contrast, channel concatenation can lead to information congestion, making motion signals less effective. We find that initializing the normalization layer as the identity function is vital for optimal performance. Additionally, placing the MGF module within the Temporal DiT block significantly enhances trajectory motion control, evidenced by a drop in Trajectory Error from 23.39 to 14.25. 

\begin{table}[!t]\footnotesize
\centering
\begin{tabular}{ccccc}
\toprule
Motion-guidance            & FVD~($\downarrow$) & CLIPSIM~($\uparrow$) & TrajError~($\downarrow$) \\ 
\midrule
Dense Flow &   601  &      0.2307   &               39.34   \\
Sparse Flow  &  556   &    0.2334    &        24.73          \\
Hybrid             &  \textbf{513}   &  \textbf{0.2358}       &   \textbf{14.25}               \\ 
\bottomrule
\end{tabular}
\caption{Ablation of the type of training trajectories. ``Hybrid" denotes the two-stage training strategy.}
\label{tab4}
\vspace{-3mm}
\end{table}

\noindent \textbf{Training Strategies.} We evaluate the two-stage training approach, with results in Table~\ref{tab4}. Training only with dense optical flow is ineffective, as it fails to capture the details of sparse trajectories. Conversely, using only sparse trajectories provides limited information, complicating the learning process. In contrast, our two-stage strategy demonstrates better adaptability and versatility in managing various motion patterns, leading to improved overall performance.

\begin{figure}[!t]
    \centering
    \includegraphics[width=1\linewidth]{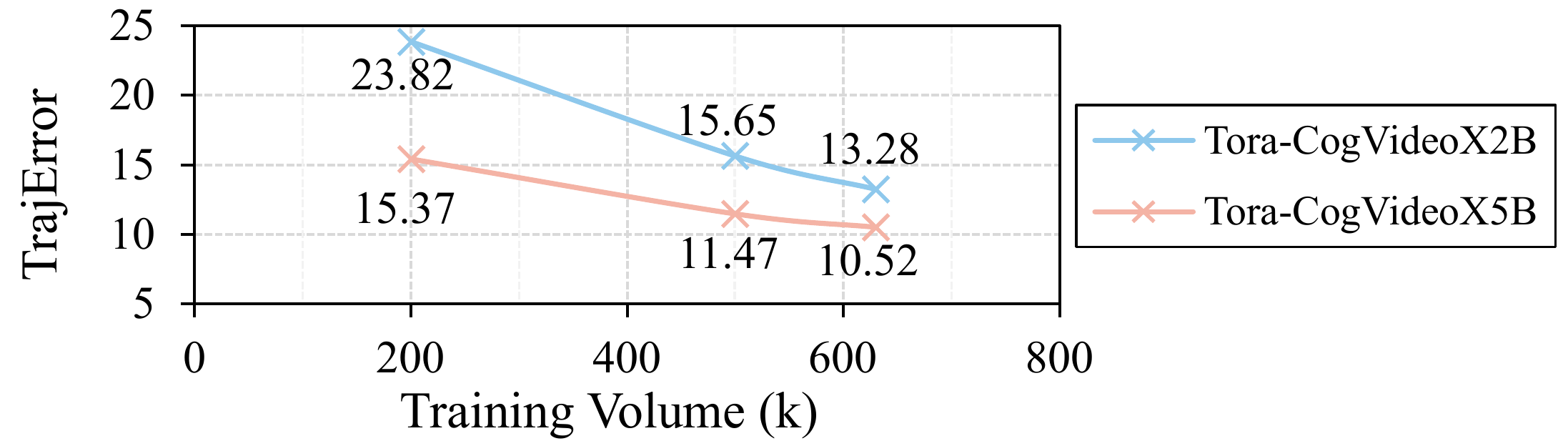}
    \caption{Scaling behavior of motion control ability in Tora.}
    \label{f1}
    \vspace{-4mm}
\end{figure}

\noindent \textbf{Scaling motion-control ability.} We investigate scaling laws in motion-controllable video generation by examining model parameter size and training volume. To leverage existing multi-size foundational models, we transfer our motion modules to CogVideoX architectures (2B/5B parameters), resulting in Tora-CogVideoX2B (2.5B) and Tora-CogVideoX5B (6.3B). Our implementation retains base VAE for motion compression while inserting MGF prior to Expert Transformer's full attention module. Figure~\ref{f1} demonstrates that increasing model scale and training data improves motion control well, confirming our modules' seamless compatibility with DiT's scalability framework.

\section{Conclusion}
This paper introduces Tora, the first trajectory-oriented Diffusion Transformer framework for video generation. 
Tora effectively encodes arbitrary trajectories into spacetime motion patches, which align well with the scaling properties of DiT, thereby enabling more realistic simulations of physical world movements. By employing a two-stage training process, Tora achieves motion-controllable video generation across a wide range of durations, aspect ratios, and resolutions. Remarkably, it can generate high-quality videos that adhere to specified trajectories, producing up to 204 frames at 720p resolution. This capability underscores Tora's versatility and robustness in handling diverse motion patterns while maintaining high visual fidelity. We hope our work provides a strong baseline for future research in motion-guided Diffusion Transformer methods.


\cleardoublepage
\vspace{6mm}
\begin{center}
   {\huge \textbf{Appendix}}
\end{center}

This supplementary material offers additional results, comprehensive dataset information, and thorough analyses that bolster the findings and conclusions outlined in the main text. It is organized as follows:

\begin{itemize}
\item Additional qualitative results. 
\item User Study with DiT-based methods.
\item Data pre-processing method. 
\item Dataset details, regarding total quantity, total durations, etc. 
\item Prompt refinement method. 
\item Motion VAE training.
\end{itemize}

\section{Qualitative Comparisons}\label{sup.results}

While the main text focuses on quantitative comparisons with the motion-controllable video generation models and ablation studies on different designs for Trajectory Extractor and Motion-guidance Fuser, here we provide further visual comparisons.


\begin{figure*}[!ht]
    \centering
    \includegraphics[width=\textwidth]{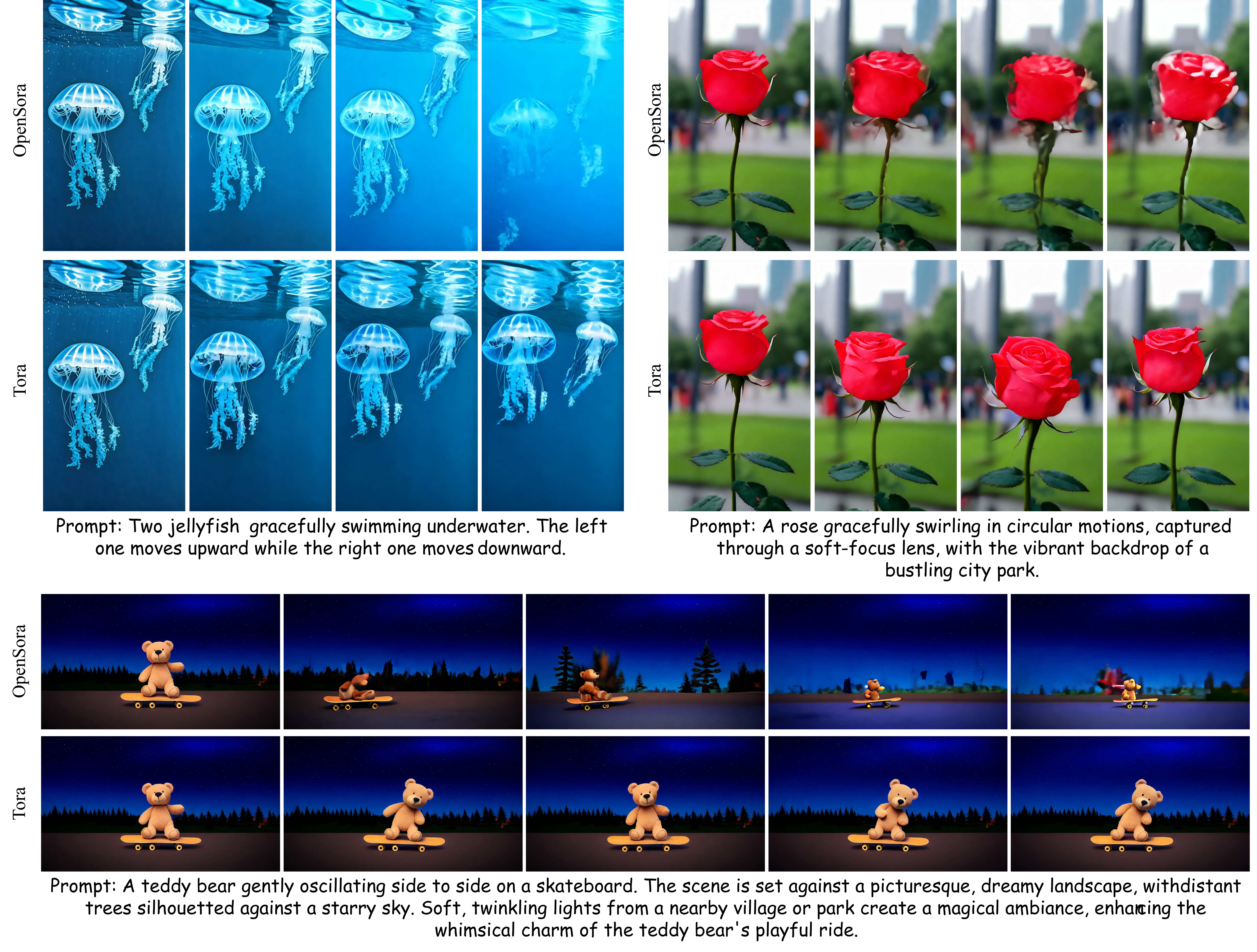}
    \caption{
        Qualitative comparison between Tora and OpenSora. All results are generated under the same text and image conditions. Tora employs an appropriate trajectory that simulates real-world physics, leading to more coherent and stable motion.
    }
    \label{sup.f.opensora}   
\end{figure*}


\subsection{Compare with OpenSora}

Despite OpenSora's impressive accomplishments, it faces challenges when creating long videos featuring complex motions, such as simultaneous movement of multiple objects, swinging, or circling. This often leads to incoherent or distorted foreground objects, negatively impacting visual quality. To our delight, we discovered that incorporating appropriate trajectory control into the DiT model offers a more effective constraining signal. This improvement markedly enhances video fluidity and preserves object fidelity, as demonstrated in Figure ~\ref{sup.f.opensora}.

\begin{figure*}[!ht]
    \centering
    \includegraphics[width=\textwidth]{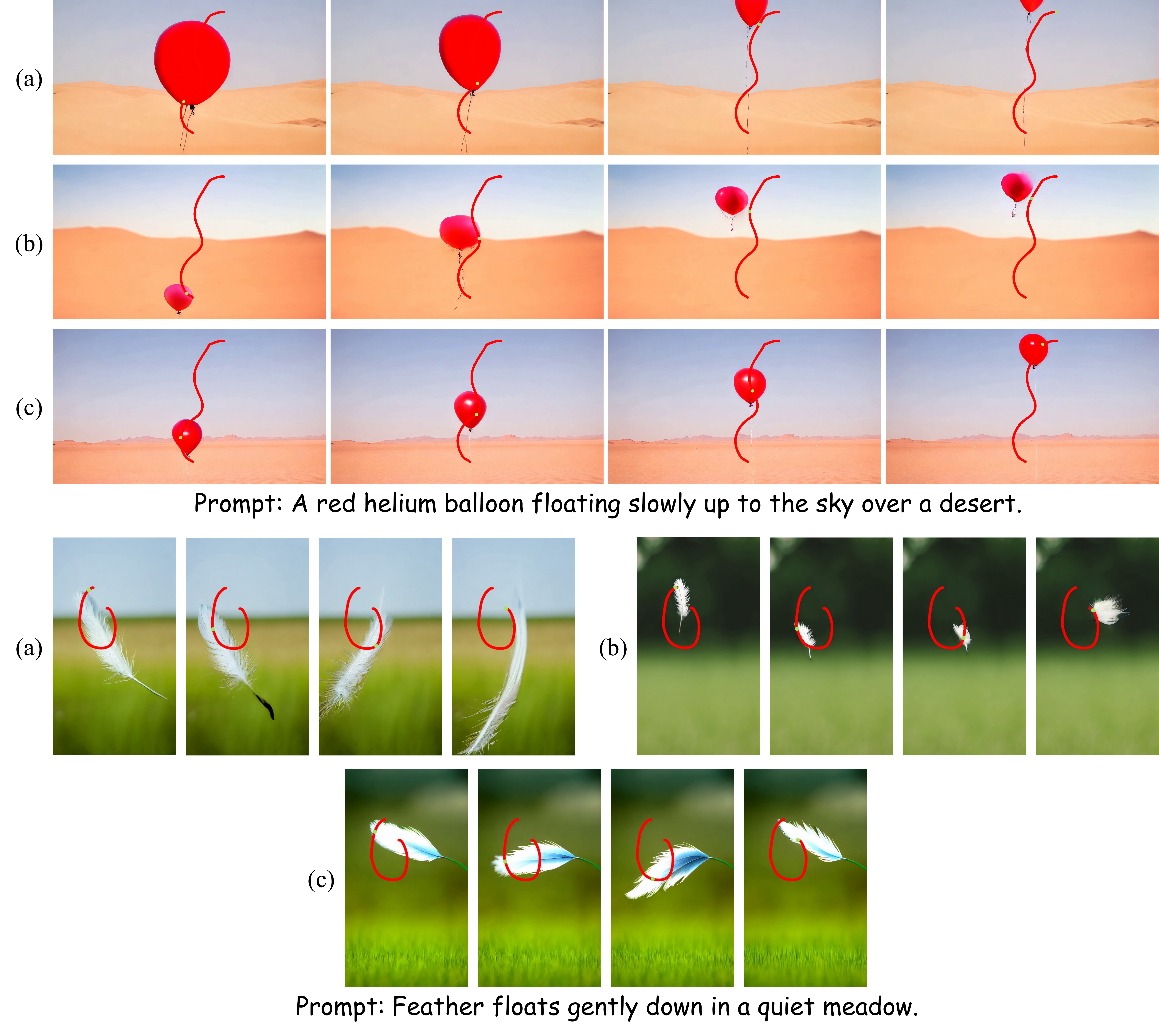}
    \caption{
        Generated videos employing different trajectory compression methods: (a) Sampling Keyframe; (b) Average Pooling; (c) 3D VAE.
    }
    \label{sup.f.compress}   
\end{figure*}

In scenarios where a teddy bear is oscillating side to side on a skateboard or a rose is swirling in circular motions, OpenSora, which relies solely on textual directives for motion control, exhibits noticeable object deformations. In contrast, Tora excels at maintaining the inherent shape of the objects. Additionally, when managing the motion of multiple entities, such as a pair of jellyfish—one moving upward while the other moves downward, OpenSora demonstrates noticeable flickering, underscoring its limitations in handling complex movements. In conclusion, the integration of Tora's motion signaling mechanism enhances both the controllability and stability of the synthesized video output.

\subsection{Comparison of Different Trajectory Compression Methods}
We train our proposed trajectory extractor using the various trajectory compression methodologies previously discussed. The comparisons of these methods are visually illustrated in Figure \ref{sup.f.compress}.

\begin{figure*}[!ht]
    \centering
    \begin{subfigure}[c]{0.3\textwidth}
        \centering
        \includegraphics[width=\linewidth]{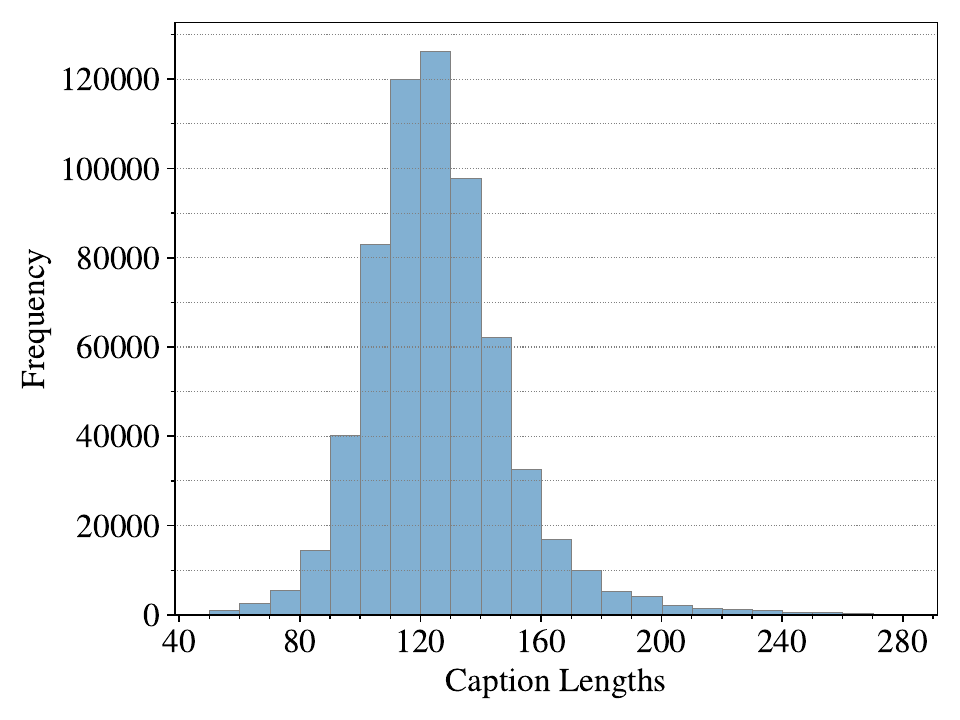}
        \caption{Histogram of Caption Lengths.}
        \label{sup.f.hist1}
    \end{subfigure}%
    \hfill
    \begin{subfigure}[c]{0.3\textwidth}
        \centering
        \includegraphics[width=\linewidth]{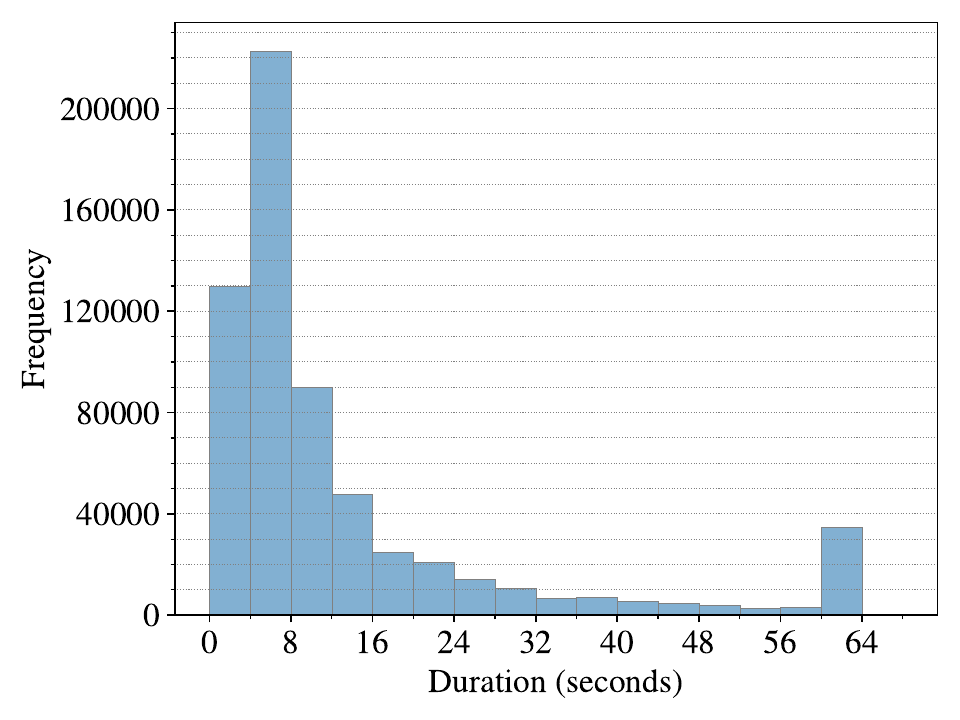}
        \caption{Histogram of Video Durations.}
        \label{sup.f.hist2}  
    \end{subfigure}%
    \hfill
    \begin{subfigure}[c]{0.3\textwidth}
        \centering
        \includegraphics[width=\linewidth]{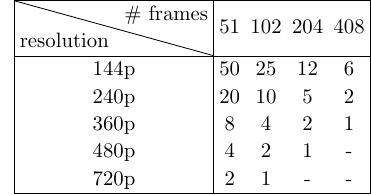}
        \caption{The training batch size of every bucket (resolution, duration).}
        \label{sup.f.bucket}   
    \end{subfigure}
    
    \caption{Overview of the training data distributions and batch sizes.}
    \label{fig:combined}
\end{figure*}

In key-frame sampling, while it successfully captures essential motion, it frequently leads to misalignment between video patches and motion patches, especially during rapid motion sequences. This misalignment hinders the generated objects from accurately tracking their trajectories, negatively impacting visual fluidity and overall quality. On the other hand, average pooling smooths out minor variations, resulting in a more consistent motion representation. However, in complex trajectories, such as S-shaped turns where consecutive frame directions are inconsistent, this approach may introduce artifacts because the physical relevance of optical flow decreases. In contrast, our proposed 3D VAE approach effectively compresses trajectory information into the video's latent space. By training the 3D VAE on the large dataset with flow annotations, it successfully extracts the most relevant motion features for guidance, preserving the movement of successive frames to a significant extent. As evidenced in the results, this method significantly enhances the fluidity and coherence of the generated movements, producing visually compelling sequences that closely resemble natural motion.

\section{User Study with DiT-based methods}
We conduct a user study to compare OpenSora-v1.2, CogVideoX-2B~\cite{yang2024cogvideox}, Vidu~\cite{DBLP:journals/corr/abs-2405-04233}, and Kling v1.0~\cite{Kling}, assessing the effectiveness of our method using our evaluation dataset. 10 human volunteers participate in evaluating quality based on three criteria: physics simulation, sensory quality, and instruction adherence. For Tora, participants draw appropriate trajectories in response to given text prompts. The experiment employs a pairwise comparison approach, where evaluators choose the superior output from each pair of generated results based on the same input. The resulting win rates appear in Table \ref{sup.userstudy}. Our method outperforms both OpenSora and CogVideoX-2B across all metrics, affirming the superiority of our proposed modules and data processing methods. Compared to the closed-source method, Vidu, we achieve competitive results. Kling demonstrates remarkable capabilities, and we hope that Tora can work to close the performance gap in future iterations. 

\begin{table}[!ht]
\centering
\fontsize{8.0pt}{9.5pt}\selectfont
\begin{tabular}{cccc}
\toprule
Model          & Phys. Simu.  & Sens. Qual.  & Inst. Foll. \\ 
\midrule
Tora vs. OpenSora-v1.2 & 71\% & 61\% & 64\%     \\
Tora vs. CogVideoX  & 53\% & 56\% & 52\%     \\ 
Tora vs. Vidu  & 54\% & 48\% & 47\%     \\ 
Tora vs. Kling  & 45\% & 43\% & 41\%     \\ 
\bottomrule
\end{tabular}
\caption{Win rates of Tora compared to OpenSora-v1.2, CogVideoX, Vidu, and Kling in terms of Physics Simulation, Sensory Quality, and Instruction Following.}
\label{sup.userstudy} 
\end{table}

\section{Data Pre-processing}\label{sup.data-proc}

During the processing of the video datasets, constructing a high-quality training set is crucial as it significantly impacts the quality of the generated videos. The following is a detailed description of our data processing workflow, which includes steps such as invalid videos removal, resolution filtering, camera motion filtering, and assessing the degree of object motion.

Initially, during the dataset preparation phase, we remove invalid videos. This step aims to identify and discard videos that do not meet our established criteria, including those with encoding errors, a duration of zero, or low quality. We identify encoding errors and zero-duration videos by directly decoding them. Furthermore, we predict both the aesthetic score\footnote{https://github.com/christophschuhmann/improved-aesthetic-predictor} and the optical flow score~\cite{DBLP:journals/pami/XuZCRYTG23} for each video. A video is deemed valid only if its aesthetic score exceeds 5.5 and its flow score is greater than 3.

Next, we perform resolution filtering. To ensure the effectiveness of subsequent study, we establish a minimum resolution standard of 720p. By checking the resolution of each video, we can eliminate those that fall below this threshold, thereby ensuring that the videos in our dataset possess adequate clarity and detail.

Subsequently, we perform camera motion filtering using a camera motion detector\footnote{https://github.com/antiboredom/camera-motion-detector} and a motion segmentor ~\cite{DBLP:conf/eccv/ZhaoLGWL22} to filter out videos with significant camera movement, which may distort the model's ability to focus on the motion of the primary subjects. More specifically, the zoom detection threshold is set between 0.4 and 0.6. The detected camera movement angles, calculated based on the background from the motion segmentation results, are valid as follows:$[0^{\circ}, 20^{\circ}], [160^{\circ}, 200^{\circ}], [340^{\circ}, 360^{\circ}]$. 

Finally, we analyze the magnitude score of the optical flow within the foreground, excluding those scenes that are mostly static or exhibit minimal movement. Moreover, dramatic object motions in some videos can cause significant optical flow deviations, interfering with trajectory training. Consequently, we retain these videos with a probability of $(1 - flow\_score / 100)$. 

Through these rigorous filtering and processing steps, we successfully construct a high-quality video dataset suitable for subsequent training, providing a solid foundation for our study.

\section{Dataset Details}\label{sup.data-detail}

This section offers an overview of the dataset used in this study, covering its origin and composition. We employ histograms and descriptive statistics to illustrate the dataset's structure and distribution.

\subsection{Training Data}

The video data is sourced from the Panda-70M subset, Mixart, and internal videos. We initially collect 2.6M videos and apply the data preprocessing pipeline to filter the content, resulting in 631k eligible videos for training. An overview of the training dataset is presented in Table \ref{sup.t1}, which details the durations, resolution and FPS.

\begin{table}[!ht]
\centering
\renewcommand{\arraystretch}{1.1}
\begin{tabular}{lr}
\toprule
    \# Videos Clips &  631053 \\
    Total Durations (hours) &  2952.93  \\
    Average Shorter Edge Length  &  965.11 \\
    Average FPS   &  29.23     \\ 
\bottomrule
\end{tabular}
\caption{Statistical information about the training data.}
\label{sup.t1}
\end{table}

Additionally, Table \ref{sup.t2} summarizes the mean and standard deviation for the durations, number of frames, and caption lengths. We also present histogram to show the distribution of the caption lengths and the durations of all video clips, as shown in the Figure \ref{sup.f.hist1} and Figure \ref{sup.f.hist2}. 

\begin{table}[!ht]
\centering
\renewcommand{\arraystretch}{1.1}
\begin{tabular}{lrr}
\toprule
    & mean & std \\ 
\midrule
    Durations (seconds) &  16.85  &    19.58    \\
    \#Frames &  506.22   &  644.38 \\
    Caption Length (\#word)   &  125.52 &  24.22  \\ 
\bottomrule
\end{tabular}
\caption{Statistics of training set, regarding durations, number of frames, and caption lengths.}
\label{sup.t2}
\end{table}

\begin{figure*}[!ht]
    \centering
    \includegraphics[width=\textwidth]{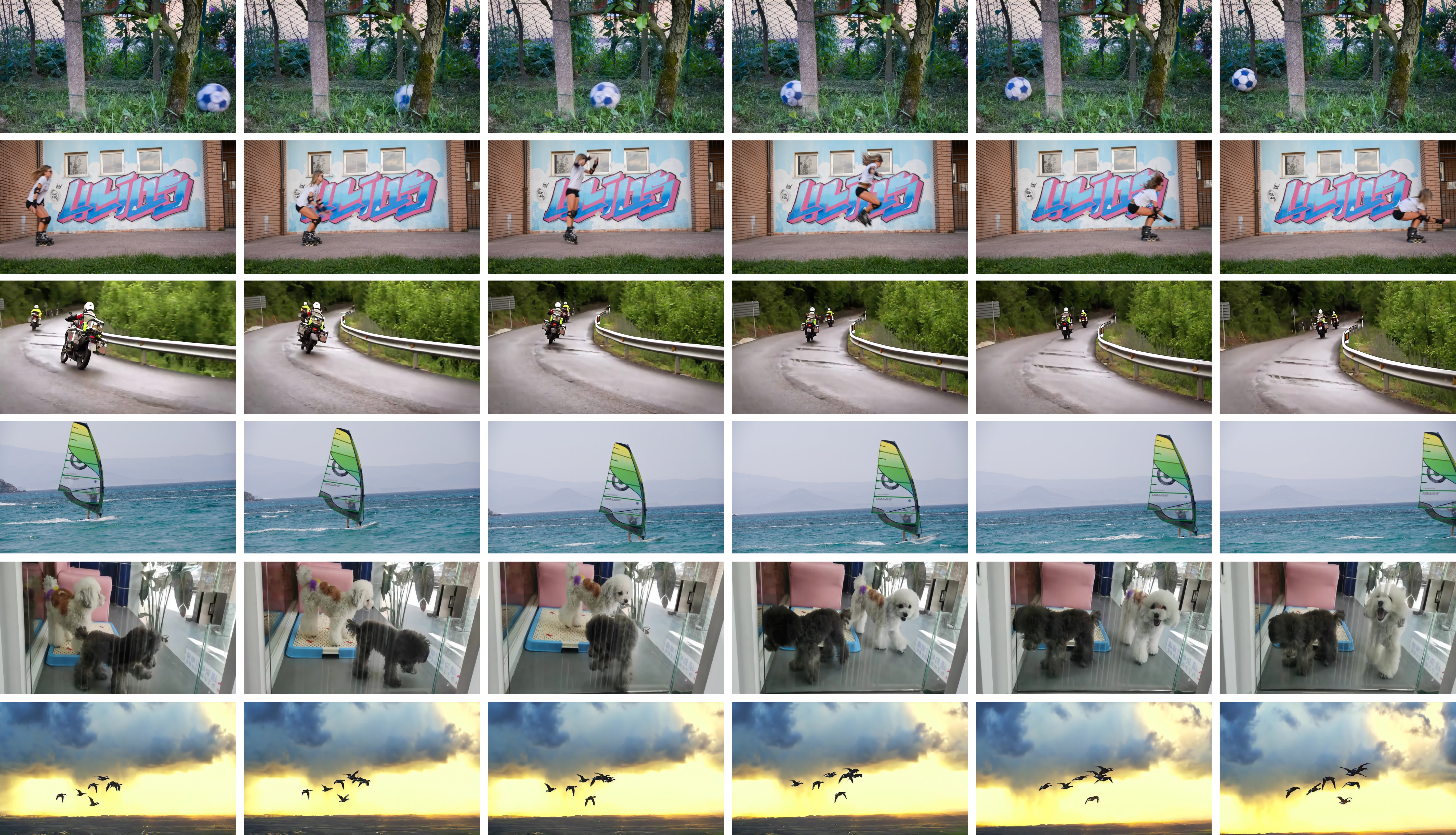}
    \caption{
       Visualization of Our Evaluation Dataset, highlighting 0\%, 20\%, 40\%, 60\%, 80\%, and 100\% of the total duration. Each center point of the annotated object masks is treated as a trajectory point. The number of trajectories in the tested video matches the number of annotated objects.
    }
    \label{sup.f.eval}   
\end{figure*}

Drawing inspiration from OpenSora, we employ a multi-scale and mixed-duration training strategy, which involves training videos of various resolutions and lengths together. Specifically,we establish predefined buckets, each defined by a unique combination of (video resolution, duration). Videos are then assigned to the appropriate bucket according to their specific attributes. Note that videos of any aspect ratio will fall into these buckets if their total pixel count is within the specified statistical intervals. The parameter settings for the buckets adhere to the principle that lower resolutions correspond to longer durations, enabling Tora to adapt to videos of varying lengths. Notably, our preprocessing steps ensure that the shorter edge of each training video exceeds 720 pixels. To enable training across various scales, we shuffle the dataset and randomly select videos for downsampling to lower resolutions. Additionally, we employ different batch sizes for each bucket to balance the GPU load. The details of the buckets are presented in Figure~\ref{sup.f.bucket}.

\subsection{Evaluation Data}

Our evaluation dataset is primarily sourced from video object segmentation datasets~\cite{yt2021,qi2022occluded,pont2017}, which offer robust object motion critical for our analysis. To enhance the quality of our evaluation, we implement a camera motion filtering technique to select videos where the camera remains predominantly stable. This filtering process allows us to concentrate on where object motion is distinctly pronounced, thereby improving the reliability of our assessments. For each frame, we utilize the center of the annotated object masks as trajectory points, providing precise references for evaluating motion dynamics. Figure~\ref{sup.f.eval} presents several examples from our evaluation dataset, highlighting the diversity and relevance of the selected video sequences.


\section{Prompt Refinement}\label{sup.prompt}
We encourage users to provide detailed text prompts to achieve satisfactory video results. To ensure consistency in the distribution of text prompts during both training and testing phases, we utilize GPT-4o to refine simple testing prompts. The process of learning refined prompts for GPT-4o involves two key components. The first component is the task description, which clearly outlines the objectives for the model in generating enhanced content:

\texttt{You need to refine user's input prompt. The user's input prompt is used for video generation task. You need to refine the user's prompt to make it more suitable for the task. Here are some examples of refined prompts: $\downarrow$ \\ 
a close-up shot of a woman applying makeup. she is using a black brush to apply a dark powder to her face. the woman has blonde hair and is wearing a black top. the background is black, which contrasts with her skin tone and the makeup. the focus is on her face and the brush, with the rest of her body and the background being out of focus. the lighting is soft and even, highlighting the texture of the makeup and the woman's skin. there are no texts or other objects in the video. the woman's expression is neutral, and she is looking directly at the camera. the video does not contain any action, as it is a still shot of a woman applying makeup. the relative position of the woman and the brush is such that the brush is in her hand and is being used to apply the makeup to her face. the video does not contain any other objects or actions. the woman is the only person in the video, and she is the main subject. the video does not contain any sound. the description is based on the visible content of the video and does not include any assumptions or interpretations. $\downarrow$ \\
a professional setting where a woman is presenting a slide from a presentation. she is standing in front of a projector screen, which displays a bar chart. the chart is colorful, with bars of different heights, indicating some sort of data comparison. the woman is holding a pointer, which she uses to highlight specific parts of the chart. she is dressed in a white blouse and black pants, and her hair is styled in a bun. the room has a modern design, with a sleek black floor and a white ceiling. the lighting is bright, illuminating the woman and the projector screen. the focus of the image is on the woman and the projector screen, with the background being out of focus. there are no texts visible in the image. the relative positions of the objects suggest that the woman is the main subject of the image, and the projector screen is the object of her attention. the image does not provide any information about the content of the presentation or the context of the meeting. $\downarrow$ \\
a serene scene in a park. the sun is shining brightly, casting a warm glow on the lush green trees and the grassy field. the camera is positioned low, looking up at the towering trees, which are the main focus of the image. the trees are dense and full of leaves, creating a canopy of green that fills the frame. the sunlight filters through the leaves, creating a beautiful pattern of light and shadow on the ground. the overall atmosphere of the video is peaceful and tranquil, evoking a sense of calm and relaxation. $\downarrow$ \\
a moment in a movie theater. a couple is seated in the middle of the theater, engrossed in the movie they are watching. the man is dressed in a casual outfit, complete with a pair of sunglasses, while the woman is wearing a cozy sweater. they are seated on a red theater seat, which stands out against the dark surroundings. the theater itself is dimly lit, with the screen displaying the movie they are watching. the couple appears to be enjoying the movie, their attention completely absorbed by the on-screen action. the theater is mostly empty, with only a few other seats visible in the background. the video does not contain any text or additional objects. the relative positions of the objects are such that the couple is in the foreground, while the screen and the other seats are in the background. the focus of the video is clearly on the couple and their shared experience of watching a movie in a theater. $\downarrow$ \\
a scene where a person is examining a dog. the person is wearing a blue shirt with the word "volunteer" printed on it. the dog is lying on its side, and the person is using a stethoscope to listen to the dog's heartbeat. the dog appears to be a golden retriever and is looking directly at the camera. the background is blurred, but it seems to be an indoor setting with a white wall. the person's focus is on the dog, and they seem to be checking its health. the dog's expression is calm, and it seems to be comfortable with the person's touch. the overall atmosphere of the video is calm and professional. $\downarrow$ \\
The refined prompt should pay attention to all objects in the video. The description should be useful for AI to re-generate the video. The description should be no more than six sentences. The refined prompt should be in English.}

Following that, GPT-4o is supplied with the testing captions for processing. This allows it to refine the prompts based on the initial task description, ensuring that the provided captions are more detailed and aligned with our objectives:

\texttt{Generate the refined prompts for following inputs: $\downarrow$ \\
A man rides on a huge fish, flying from the water into the sky. $\downarrow$ \\
Two Jedi cats are fighting with each other in the forest. $\downarrow$  \\
A polar bear with a black hat is walking on the Great Wall. $\downarrow$  \\
A woman and a golden retriever are playing on the beach at sunset. $\downarrow$ \\
Two roses sway together before a snow-covered mountain range.
}

\section{Motion VAE Training}
Given the absence of pre-existing networks tailored for video optical flow compression, training such a network from scratch presents significant challenges. Directly transferring the motion vectors to the image domain and applying a pretrained 3D VAE may hinder the model's ability to effectively encode motion features, primarily due to domain discrepancies. To overcome this issue, we refine a motion-specific 3D VAE that is initialized from a pretrained model. Specifically,  our motion 3D VAE is specifically initialized using the architecture of OpenSora's VAE, which adapts the structure of Magvit-v2. This VAE has a substantial parameter count of 384 million, effectively leveraging the capabilities of a well-established network.  Our training data is sourced from a combination of datasets annotated with optical flow information~\cite{DBLP:conf/cvpr/MehlSJNB23,DBLP:conf/cvpr/MayerIHFCDB16,DBLP:journals/ijcv/RanjanHTTRB20,DBLP:journals/corr/abs-2001-10773}. We fine-tune the motion 3D VAE for 200,000 iterations with a batch size of 1. The training video size is set to a random number of frames, capped at 34. This setting aligns with the OpenSora video VAE, improving compatibility between the motion VAE and the video VAE and ensuring a cohesive training process. We utilize PSNR, SSIM and Trajectory Error to evaluate reconstruction quality and motion-controllable ability. The performance differences between the pure video VAE and our fine-tuned model are presented in Table~\ref{sup.t.vae}.

\begin{table}[!ht]
\centering
\begin{tabular}{cccc}
\toprule
Model          & PSNR$\uparrow$  & SSIM$\uparrow$  & TrajError$\downarrow$ \\ 
\midrule
Pure Video VAE & 27.34 & 0.842 & 17.09     \\
Our VAE        & 28.76 & 0.860 & 14.25     \\ 
\bottomrule
\end{tabular}
\caption{The performance comparison of different 3D VAE.}
\label{sup.t.vae} 
\end{table}

{
    \small
    \bibliographystyle{ieeenat_fullname}
    \bibliography{main}
}


\end{document}